\documentclass{article}

\usepackage{graphicx, xcolor}
\usepackage{fontenc}
\usepackage{lmodern}
\usepackage{textcomp}
\usepackage{latexsym}
\usepackage[fleqn]{amsmath}
\usepackage{amssymb}
\usepackage{amsthm}
\usepackage{bm}
\usepackage{latexsym}
\usepackage{amsfonts}
\usepackage{algorithmic}
\usepackage{algorithm}

\theoremstyle{definition}

\newtheorem*{theorem*}{Theorem}

\newtheorem*{definition*}{Definition}

\newtheorem*{proposion*}{Proposition}

\newtheorem*{example*}{Example}

\newcommand{\diag}{\mathop{\mathrm{diag}}\nolimits}

\makeatother
\DeclareMathOperator*{\argmax}{arg\, max}
\DeclareMathOperator*{\argmin}{arg\, min}

\title{A Neural Network model with Bidirectional Whitening}

\author
{Yuki Fujimoto$^{\ast}$ and Toru Ohira$^{\ast\ast}$ \\
\\
\normalsize{Graduate School of Mathematics,  Nagoya University,  Nagoya,  Japan}\\
\
\normalsize{$^\ast$E-mail:  m15042x@math.nagoya-u.ac.jp}
\normalsize{$^{\ast\ast}$E-mail:  ohira@math.nagoya-u.ac.jp}
}

\date{}

\begin{document}

\maketitle

\begin{abstract}
We present here a new model and algorithm which performs an efficient  Natural gradient descent for Multilayer Perceptrons.  Natural gradient descent was originally proposed from a point of view of information geometry,  and it performs the steepest descent updates on manifolds in a Riemannian space.  In particular,  we extend an approach taken by the ``Whitened neural networks''  model.  We make the whitening process not only in feed-forward direction as in the original model,  but also in the back-propagation phase.  Its efficacy is shown by an application of this ``Bidirectional whitened neural networks'' model to a handwritten character recognition data (MNIST data).  
\end{abstract}

\section{Introduction}

Interests for developing and efficient learning algorithm for multilayer neural networks have grown rapidly due to recent upheaval of the deep learning and other machine learnings.  Natural gradient descent(NGD) is considered as one of the strong methods.  It was proposed from a point of view of information geometry\cite{amari1998natural},  where neural networks are considered as manifolds in a Riemannian space with a measure given by the
Fisher information matrix (FIM). Then,  the learning process can be interpreted as an optimization problem of a function in a Riemannian space.  The idea of applying the NGD to multilayer neural networks was initiated by Amari.  Recently,  it has regained interests from machine learning researchers\cite{martens2014new,pascanu2013revisiting}.

However,  difficulty exists for using the NGD: the computational costs of estimating the FIM and obtaining its inverse is high.  Much attention and research efforts have gone into solving this difficulty\cite{NIPS2015_5953,martens2015optimizing,park2000adaptive,ioffe2015batch,NIPS2016_6114}. 

In this paper,  we will focus on one of such approaches,  and extend the work of \cite{NIPS2015_5953}.  In their approach ``Whitened neural networks'' model was proposed.  There,  a neural network architecture, whose FIM is closer to the identity matrix with less computational demands,  is explored.  Extra neurons and connections are added to achieve this whitening approximation.  In particular,  they have
used this scheme for the forward direction of inputs to neurons and achieved lower computational costs. 

Our main proposal in this paper is to further push the approximation of the FIM being closer to
the identity by implementing the whitening process also in the back-propagation phase. 
This model,  which we term as the ``bidirectional whitened neural networks'' model,  will
be described in the following.  Its efficacy is also shown through its application to a handwritten character recognition data (MNIST data). 

\section{Multilayer Perceptron and Natural Gradient Descent}

We present here a brief review of the Multilayer Perceptron and the Natural Gradient Descent,  which we focus in this paper.  The first level of approximation for the FIM
is also discussed. 

\subsection{Multilayer Perceptron}
\label{sec: multilayer perceptron}

Multilayer Perceptron is a model of neural networks which has feed-forward structure with no recurrent loops.  They have multiple layers called input,  hidden,  and output,  and
neurons have  all to all connections between successive layers.   Let us consider a
$N$ layer Perceptron,  and set the values  
of the input as $\bm{z}^{(0)} = \bm{x}$, the hidden layer values as $\bm{z}^{(i)} = \bm{h}^{(i)}$, ($1 \leq i \leq N - 1$), 
and the output of the entire network as $\bm{z}^{(N)} = f(\bm{x}; \bm{w})$.  

This $f(\bm{x}; \bm{w})$ can be viewed as a function of $\bm{x}$ by fixing the parameters $\bm{w}$,  and thus called as a ``multilayer Perceptron function''. The rules of computing
the value of the $i$ layer from the $i - 1$ in the network is given as follows
($1 \leq i \leq N$).
\begin{eqnarray}
\label{eq: forward propagation1}
\bm{a}^{(i)} &=& W^{(i)}\bm{z}^{(i-1)} + \bm{b}^{(i)} \\
\label{eq: forward propagation2}
&=& \bar{W}^{(i)}\bar{\bm{z}}^{(i-1)} \\
\label{eq: forward propagation3}
\bm{z}^{(i)} &=& \bm{\phi}^{(i)}(\bm{a}^{(i)})
\end{eqnarray}
Here,  $\bm{\phi}^{(i)}(\cdot)$is an activation function applied to each element of $\bm{a}$. 
Typically,  the sigmoid function or $ReLU$ function are used for this activation function. 
Also,  \eqref{eq: forward propagation2} is a shortened notation by setting
$\bar{W}^{(i)} \equiv (\bm{b}^{(i)},  W^{(i)}),  \bar{\bm{z}}^{(i)} \equiv (1,  \bm{z}^{(i)^{T}})^{T}$. 

Hence,  the multilayer Perceptron function (MPF) is defined by setting 

\noindent
$\{(W^{(i)}, \bm{b}^{(i)})\}$.  It is often convenient to denote these parameters by $\bm{w}$,  defined by
\begin{eqnarray}
\label{eq: vectolization}
\bm{w} \equiv (\mathrm{vec}(\bar{W}^{(1)})^{T},  \ldots,  \mathrm{vec}(\bar{W}^{(N)})^{T})^{T}
\end{eqnarray}
where $\mathrm{vec}(A)$ means a compound vector of column vectors of a matrix $A$

The learning process of multilayer Perceptrons is an optimization problem set by the 
following statistical inference.  The training data of input and output pairs is
given as $D \equiv \{(\bm{x}_{k},  \bm{y}_{k})\}_{k=1}^{K}$.  We assume this data set 
is generated by the same joint distribution $Q(X,  Y)$ independently.  In order to estimate
this input output probabilistic relations,  a statistical model $\{p(\bm{x},  \bm{y}; \bm{w})\}_{\bm{w} \in \Theta}$ is considered using the MPF.  Here $p(\bm{x},  \bm{y}; \bm{w})$ is 
a joint probability density function and $\Theta \subset \mathbb{R}^{M}$ is a set of parameters.  The problem is to find the parameter $\bm{w}$ which makes $p(\bm{x},  \bm{y}; \bm{w})$ as a best estimate of $Q(X,  Y)$.  The maximum likelihood	method is employed to
obtain such $\bm{w^*}$. 
\begin{eqnarray}
\bm{w}^{\ast} &\equiv& \argmax_{\bm{w} \in \Theta} \prod_{k=1}^{K} p(\bm{x}_{k},  \bm{y}_{k}; \bm{w})
\end{eqnarray}
It is known that this estimation is the same as the following minimization problem. 
\begin{eqnarray}
\label{eq: MLP Learning1}
\bm{w}^{\ast} &\equiv& \argmin_{\bm{w} \in \Theta} \sum_{k=1}^{K} -\log p(\bm{x}_{k},  \bm{y}_{k}; \bm{w}) \\
\label{eq: MLP Learning3}
&=& \argmin_{\bm{w} \in \Theta} M(\bm{w})
\end{eqnarray}
Here,  we have set the target function to minimize as $M(\bm{w})$.  Research on efficient algorithms for this optimization problem is the central issue in the following.

\subsection{Natural Gradient Method}
\label{sec: natural gradient}
Natural Gradient Method is a steepest descent method in a Riemannian space.  It is proposed from the 
information geometry where statistical models are manifolds in a Riemannian space with a metric of the Fisher Information Matrices\cite{amari2007methods}. Thus,  we can view the learning by the multilayer Perceptrons as an optimization problem in a Riemannian space as presented in \ref{sec: multilayer perceptron}.

Let us start by defining the Fisher information matrix and the Natural Gradient Descent.
\vspace{1em}

\noindent
{\textbf{Definition: Fisher Information Matrix}}
\vspace{1em}

We set $l(\bm{x} ; \bm{w}) \equiv \log p(\bm{x};\bm{w})$. 
For $\bm{w} \in \Theta$, a square matrix $G(\bm{w}) = (g_{ij} (\bm{w}))$ is defined as follows. 
\begin{eqnarray}
\label{eq:fisher information matrix}
G(\bm{w}) \equiv E \left[\nabla l(X; \bm{w}) \nabla l(X; \bm{w})^{T}\right]
\end{eqnarray}
\eqref{eq:fisher information matrix} can be expressed by each elements as,
\begin{eqnarray}
g_{ij}(\bm{w}) = E \left[ \dfrac{\partial l}{\partial w_{i}} (X ; \bm{w}) \dfrac{\partial l}{\partial w_{j}} (X ; \bm{w}) \right] 
= \int \dfrac{\partial l}{\partial w_{i}} (\bm{x} ; \bm{w}) \dfrac{\partial l}{\partial w_{j}} (\bm{x};\bm{w}) p(\bm{x}; \bm{w})d\bm{x}
\end{eqnarray}
We call this matrix $G$ the Fisher information matrix (FIM).
\clearpage

\noindent
{\bf{Definition: Natural Gradient Descent}}
\vspace{1em}

We call the following 
gradient method as  the Natural Gradient Descent (NGD).
\begin{eqnarray}
\label{eq: natural gradient descent}
\bm{w}(t + 1) \!=\! \bm{w}(t) \!-\! \eta(t) G^{-1}(\bm{w}(t)) \nabla M(\bm{w}(t))
\end{eqnarray}
Here $\eta(t)$ is a rate of the learning.
\vspace{2em}

Then,  $-G^{-1}(\bm{w}(t)) \nabla M(\bm{w}(t))$ is the direction of the maximal decrease of the target function $M$ given a fixed step size. We note that this NGD reduces to the ordinary gradient descent,  when $G$ is the identity matrix.

\subsection{Approximation of the Fisher Information Matrix}
\label{sec: approximation of FIM}

As discussed in the previous section,  the FIM and its inverse play important roles
in the calculation in the NGD.  We,  thus,  present a preliminary approximation of the FIM
in order to lessen the computational burdens\cite{martens2015optimizing}. 

Let us first compute the FIM for the multilayer Perceptrons.  The probability density function 
associated with the multilayer Perceproton function (MPF)  is given as follows. 
\begin{eqnarray}
p(\bm{x},  \bm{y}; \bm{w}) &=& p(\bm{y} | f(\bm{x}; \bm{w})) p(\bm{x})
\end{eqnarray}
Also,  the gradient vector are written concisely as in \eqref{eq: vectolization},
\begin{eqnarray}
\dfrac{\partial l}{\partial \bm{w}} \equiv \left(\mathrm{vec}\left(\dfrac{\partial l}{\partial \bar{W}^{(1)}}\right)^{T},  \ldots,  \mathrm{vec}\left(\dfrac{\partial l}{\partial \bar{W}^{(N)}}\right)^{T}\right)^{T}
\end{eqnarray}
Then,  the FIM for the MLP is given as follows.
\begin{eqnarray}
\label{eq: neural fisher information matrix}
G(\bm{w}) &=& \begin{pmatrix}
G_{1,  1} & G_{1,  2} & \cdots & G_{1,  N} \\
G_{2,  1} & G_{2,  2} & \cdots & G_{2,  N} \\
\vdots & \vdots & \ddots & \vdots \\
G_{N,  1} & G_{N,  2} & \cdots & G_{N,  N} \\
\end{pmatrix} \\
\label{eq: neural fisher information matrix2}
G_{i,  j} &\equiv& E\left[ \mathrm{vec}\left(\dfrac{\partial l}{\partial \bar{W}^{(i)}}\right) \mathrm{vec}\left(\dfrac{\partial l}{\partial \bar{W}^{(j)}}\right)^{T} \right]
\end{eqnarray}
Hence,  the FIM for the MLP is composed of the block matrices $G_{i,  j}$. 

If we further set $\delta^{(i)}_{j} = \dfrac{\partial l}{\partial a^{(i)}_{j}}$, the following is obtained.
\begin{eqnarray}
\label{eq: loss gradient}
\dfrac{\partial l}{\partial \bar{W}^{(i)}} = \bm{\delta}^{(i)}\bar{\bm{z}}^{(i-1)^{T}}
\end{eqnarray}

By putting \eqref{eq: loss gradient} into \eqref{eq: neural fisher information matrix2}, the $G_{i, j}$ can now be expressed as
\begin{eqnarray}
G_{i,  j}
&=& E\left[
 \bar{\bm{z}}^{(i-1)} \bar{\bm{z}}^{(j-1)^{T}} \otimes \bm{\delta}^{(i)}\bm{\delta}^{(j)^{T}}
 \right]
\end{eqnarray}
(Here,  $\otimes$ is the Kronecker product. )

For the efficient computation,  it is essential to approximate this FIM.  
The preliminary approximation consists of two steps.  

The first step approximation of $G_{i, j}$ is given as $\tilde{G}_{i, j}$ which is
defined as follows. 
\begin{eqnarray}
G_{i,  j}
 \label{eq: K-FAC approximation}
 &\approx& E\left[\bar{\bm{z}}^{(i-1)}\bar{\bm{z}}^{(j-1)^{T}}\right] \otimes
 E\left[\bm{\delta}^{(i)}\bm{\delta}^{(j)^{T}}\right] \\
 &\equiv& \bar{Z}_{i-1,  j-1} \otimes D_{i,  j} \nonumber \\
 &\equiv& \tilde{G}_{i,  j} \nonumber
\end{eqnarray}
This approximation means that we are inter-changing the expectation of the Kronecker products with the 
Kronecker products of the expectations.  The matrix $\tilde{G}$,  whose elements are given by replacing 
$G_{i,  j}$ of \eqref{eq: neural fisher information matrix} with $\tilde{G}_{i,  j}$,  is the first step approximation of the FIM. 
We note that the FIM is decomposed into two parts by this approximation: $\bar{\bm{z}}^{(i-1)}$ 
(the feed-forward phase part) and $\bm{\delta}^{(i)}$ (the back-propagtaing phase part). 

We perform the second step approximation on $\tilde{G}$ to obtain $\breve{G}$. 
\begin{eqnarray}
\label{eq: block diagonalization}
\breve{G} &\equiv& \diag(\tilde{G}_{1,  1},  \tilde{G}_{2,  2},  \ldots,  \tilde{G}_{N,  N})
\end{eqnarray}
Here,  $\diag(\cdots)$ denotes a block diagonal matrix,  whose non-zero diagonals are given
by the elements. 
In other words,  $\breve{G}$ is obtained from $\tilde{G}$ by setting non-diagonal elements as the
zero matrix, 
\begin{eqnarray}
\tilde{G}_{i,  j} = O \quad (i \neq j)
\end{eqnarray}
This approximation allows us to compute the FIM layer by layer independently.

\section{Whitened Neural Networks}
\label{sec: whitening neural networks}

In this section,  we present algorithms which aim to perform Natural Gradient Descent efficiently with the approximated FIM,  $\breve{G}$. 

\subsection{Natural Gradient Descent by Whitening}
\label{sec: activation whitening}
Let us first describe Whitened Neural Networks\cite{NIPS2015_5953}.  The main idea of this method is to perform the NGD by reconfiguring the network and parameters,  so that the FIM becomes closer to the identity matrix.  When the FIM is the identity matrix,  the NGD is the same as the ordinary gradient descent, thus can be implemented simply with less computational costs.   

\subsubsection{Whitened Neural Network}
 The architecture of the Whitened Neural Networks (WNN) is obtained by changing \eqref{eq: forward propagation1} through \eqref{eq: forward propagation3} into the following form. 
\begin{eqnarray}
\label{eq: whitening forward prop1}
\bm{z}^{\dagger^{(i-1)}} &=& U^{(i-1)}(\bm{z}^{(i-1)} - \bm{c}^{(i-1)}) \\
\label{eq: whitening forward prop2}
\bm{a}^{(i)} &=& W^{\dagger^{(i)}} \bm{z}^{\dagger^{(i-1)}} + \bm{b}^{\dagger^{(i)}} \\
\label{eq: whitening forward prop3}
\bm{z}^{(i)} &=& \bm{\phi}^{(i)} \left( \bm{a}^{(i)} \right)
\end{eqnarray}
Here $\{(U^{(i-1)},  \bm{c}^{(i-1)})\}$ are the new parameters introduced as ``Whitening'' parameters.  $\{(W^{\dagger^{(i)}},  \bm{b}^{\dagger^{(i)}})\}$ are the new model parameters associated with this new architecture.  These are the ones which we want to estimate and update using gradient descent methods as in the normal multilayer Perceptrons. 

\begin{figure}[h]
\centering
\includegraphics[width=10cm]{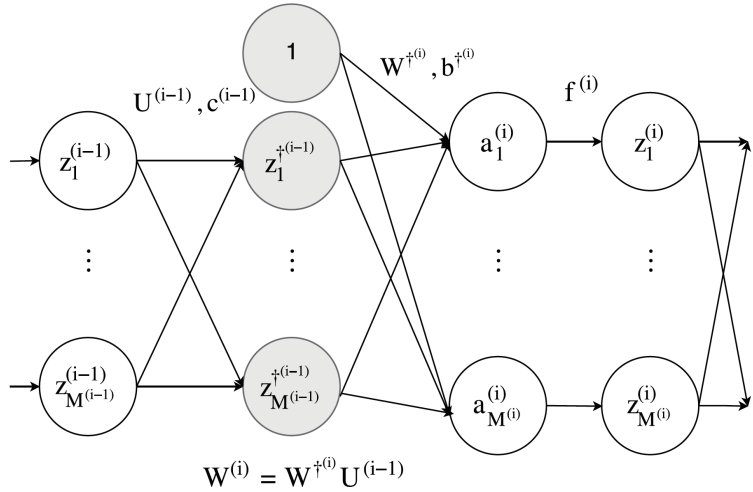}
\caption{Architecture of Whitened Neural Networks}
\label{fig:Whitening Neural Networks}
\end{figure}

We present in the Figure \ref{fig:Whitening Neural Networks} the new architecture defined by
\eqref{eq: whitening forward prop1},  \eqref{eq: whitening forward prop2},  \eqref{eq: whitening forward prop3}.  It shows the $i-1$th layer to the $i$th layer. We note the gray layer in the Figure \ref{fig:Whitening Neural Networks} is the new inserted layer for the purpose of ``whitening''. This change of network configuration is the essence of WNN. 

From \eqref{eq: K-FAC approximation}, the approximated FIM $\tilde{G}_{i,  i}$ in the WNN,  then,  is expressed as the following. 
\begin{eqnarray}
\label{eq: fisher approx}
\tilde{G}_{i,  i} = E\left[\bar{\bm{z}}^{\dagger^{(i-1)}}\bar{\bm{z}}^{\dagger^{(i-1)^{T}}}\right] \otimes
 E\left[\bm{\delta}^{(i)}\bm{\delta}^{(i)^{T}}\right]
\end{eqnarray}
The essential idea of the whitening is to make $\breve{G}$ closer to the identity by 
defining the whitening parameters $\{(U^{(i-1)},  \bm{c}^{(i-1)})\}$ as
\begin{eqnarray}
\label{eq: whitening condition}
E\left[\bar{\bm{z}}^{\dagger^{(i-1)}}\bar{\bm{z}}^{\dagger^{(i-1)^{T}}}\right] = I
\end{eqnarray}
for each $i$ and performs the gradient descent.  (Our idea,  which will be described later in \ref{sec: gradient whitening}, is to further extend the whitening to the latter factor 
$E\left[\bm{\delta}^{(i)}\bm{\delta}^{(i)^{T}}\right]$ in \eqref{eq: fisher approx})

\subsubsection{Updating of the Whitening Parameters}

We calculate here explicitly $\{(U^{(i-1)},  \bm{c}^{(i-1)})\}$,  which satisfies the condition \eqref{eq: whitening condition}. 

As $\bar{\bm{z}}^{\dagger^{(i-1)}} = (1,  \bm{z}^{\dagger^{(i-1)^{T}}})^{T}$,  \eqref{eq: whitening condition} can be decomposed into
\begin{eqnarray}
\label{eq: whitening condition2}
\begin{pmatrix}
1 & E\left[\bm{z}^{\dagger^{(i-1)^{T}}} \right] \\
E\left[\bm{z}^{\dagger^{(i-1)}}\right] & E\left[\bm{z}^{\dagger^{(i-1)}}\bm{z}^{\dagger^{(i-1)^{T}}}\right]
\end{pmatrix}
= I
\end{eqnarray}
Thus,  
\begin{eqnarray}
\label{eq: whitening condition3}
E\left[\bm{z}^{\dagger^{(i-1)}}\right] &=& \bm{0} \\
\label{eq: whitening condition4}
E\left[\bm{z}^{\dagger^{(i-1)}}\bm{z}^{\dagger^{(i-1)^{T}}}\right] &=& I
\end{eqnarray}
are required to satisfy this condition.  

Let us look at these conditions. \eqref{eq: whitening condition3} can be satisfied by
\begin{eqnarray}
\label{eq: update c}
\bm{c}^{(i-1)} &\leftarrow& E\left[\bm{z}^{(i-1)}\right]
\end{eqnarray}
Also,  for \eqref{eq: whitening condition4},  we first set the matrix $\check{Z}_{i-1,  i-1}$ by the following
\begin{eqnarray}
\check{Z}_{i-1,  i-1}
\equiv E\left[(\bm{z}^{(i-1)} \! - \! \bm{c}^{(i-1)})(\bm{z}^{(i-1)} \! - \! \bm{c}^{(i-1)})^{T}\right]
\end{eqnarray}
Then,  \eqref{eq: whitening condition4} becomes
\begin{eqnarray}
\label{eq: whitening condition6}
E\left[  \bm{z}^{\dagger^{(i-1)}} \! \bm{z}^{\dagger^{(i-1)^{T}}}  \right]
= U^{(i-1)} \check{Z}_{i-1,  i-1} U^{(i-1)^{T}} 
= I
\end{eqnarray}
Because $\check{Z}_{i-1,  i-1}$ is a symmetric matrix,  there exists a orthogonal matrix $P$,  which makes it diagonal. 
\begin{eqnarray}
\check{Z}_{i-1,  i-1}  =  P  \Lambda  P^{T}
\end{eqnarray}
Here $\Lambda$ is the diagonalized matrix.  Then,  if we set
\begin{eqnarray}
\label{eq: update U}
U^{(i-1)}  &\leftarrow& (\Lambda + \varepsilon I)^{-\frac{1}{2}} \cdot  P^{T}
\end{eqnarray}
the condition \eqref{eq: whitening condition6} is approximately satisfied. 
(Here, $\varepsilon$ is a small positive constant to avoid division by zero. )

By this process,  called the whitening process,  according to
\eqref{eq: update c} and \eqref{eq: update U},  we update the whitening parameters satisfying \eqref{eq: whitening condition}.  We note that,  in this updating,  the calculation of $\bm{z}^{(i-1)}$ in feed-forward phase is essential. 

\subsubsection{Updating of the model parameters} 

We now turn our attention to the updating of the model parameters $\{(W^{\dagger^{(i)}},  \bm{b}^{\dagger^{(i)}})\}$.  We need to pay attention so that the inclusion of the whitening process and the associated layer does not change the value of the multilayer Perceptron function (MPF) itself.  In concrete,  we need to do the following. 
Let us assume the whitening parameters $\{(U^{(i-1)},  \bm{c}^{(i-1)})\}$ are updated to $\{(U^{(i-1)}_{new},  \bm{c}^{(i-1)}_{new})\}$. 
We want to keep the value of \eqref{eq: whitening forward prop2} unchanged by this updating. 
This places a constrains in the way we update the model parameters $\{(W^{\dagger^{(i)}}_{new},  \bm{b}^{\dagger^{(i)}}_{new})\}$.  Namely,  for any value of $\bm{z}^{(i-1)}$, 
the following must be satisfied. 
\begin{eqnarray}
{W^{\dagger^{(i)}}U^{(i-1)}(\bm{z}^{(i-1)} - \bm{c}^{(i-1)}) + \bm{b}^{\dagger^{(i)}}
} = W^{\dagger^{(i)}}_{new}U^{(i-1)}_{new}(\bm{z}^{(i-1)} - \bm{c}^{(i-1)}_{new}) + \bm{b}^{\dagger^{(i)}}_{new}
\end{eqnarray}
We can obtain the following by solving these equations. 
\begin{eqnarray}
\label{eq: model param update1}
W^{\dagger^{(i)}}_{new} &=& W^{\dagger^{(i)}}U^{(i-1)}U^{(i-1)^{-1}}_{new}\\
\label{eq: model param update2}
\bm{b}^{\dagger^{(i)}}_{new} &=& \bm{b}^{\dagger^{(i)}} - W^{\dagger^{(i)}}U^{(i-1)}\bm{c}^{(i-1)} + W^{\dagger^{(i)}}_{new}U^{(i-1)}_{new}\bm{c}^{(i-1)}_{new}
\end{eqnarray}
By putting together \eqref{eq: whitening forward prop1} and \eqref{eq: whitening forward prop2},  we can set $\{(W^{(i)},  \bm{b}^{(i)})\}$ as
\begin{eqnarray}
W^{(i)} &=& W^{\dagger^{(i)}}U^{(i-1)} \\
\bm{b}^{(i)} &=& \bm{b}^{\dagger^{(i)}} - W^{\dagger^{(i)}}U^{(i-1)}\bm{c}^{(i-1)}
\end{eqnarray}
Using these $\{(W^{(i)},  \bm{b}^{(i)})\}$,  we can re-write \eqref{eq: model param update1}and \eqref{eq: model param update2} as
\begin{eqnarray}
\label{eq: update W}
W^{\dagger^{(i)}} &\leftarrow& W^{(i)}U^{(i-1)^{-1}}_{new}\\
\label{eq: update b}
\bm{b}^{\dagger^{(i)}} &\leftarrow& \bm{b}^{(i)} + W^{(i)}\bm{c}^{(i-1)}_{new}
\end{eqnarray}
Thus,  we can keep MPF the same by updating whitening parameters first as in \eqref{eq: update c} and \eqref{eq: update U} and then update model parameters with \eqref{eq: update W} and \eqref{eq: update b}. 

As we change model parameters,  the values of $E[\bm{z}^{(i-1)}],  \check{Z}_{i-1,  i-1}$ changes, which
in turn requires the update of the whitening parameters to keep the FIM close to the identity
matrix.  However,  it is computationally expensive to update both set of parameters at every iterations.  In particular, the update of the whitening parameters for a layer of $M$ neurons
takes computation of the order of $O(M^{3})$.  Thus,  in actual implementations,  the update of
the whitening parameters are performed at certain fixed time intervals\cite{NIPS2015_5953},  though this makes a gradual digression from the NGD for that time interval between the successive updating of the whitening parameters.

The method and algorithm described above is called ``Projected Natural Gradient Descent''(PRONG)\cite{NIPS2015_5953},  which is outlined in Algorithm \ref{alg: PRONG}.
\clearpage
 
\begin{algorithm}[h]
\caption{Projected Natural Gradient Descent (PRONG)\cite{NIPS2015_5953}.}
\label{alg: PRONG}
\begin{algorithmic}
\STATE{\bf Input:} training set $D$, initial parameter $\bm{w}(0)$
\STATE{\bf Hyper parameters:} updating period of whitening parameters $\tau$ 
\STATE $\bullet$ $U^{(i)} \leftarrow I; \bm{c}^{(i)} \leftarrow \bm{0}; t \leftarrow 0$ 
\WHILE{ending condition not satisfied} 
\IF{$\mathrm{mod}(t,  \tau) = 0$}
\FOR{all layers i}
\STATE $\bullet$ Computation of standard parameters $\{(W^{(i)},  \bm{b}^{(i)})\}$. 
\STATE $\bullet$ Estimations of $E[\bm{z}^{(i-1)}],  \check{Z}_{i-1,  i-1}$. 
\STATE $\bullet$ Updating of the Whitening parameters $\{(U^{(i-1)},  \bm{c}^{(i-1)})\}$. 
\STATE $\bullet$ Updating of the model parameters $\{(W^{\dagger^{(i)}},  \bm{b}^{\dagger^{(i)}})\}$. 
\ENDFOR
\ENDIF
\STATE $\bullet$ Updating of $\{(W^{\dagger^{(i)}},  \bm{b}^{\dagger^{(i)}})\}$ by the ordinary gradient descent.  
\STATE $\bullet$ $t \leftarrow t + 1$ 
\ENDWHILE
\end{algorithmic}
\end{algorithm}

\subsection{Extension of Whitening}
\label{sec: gradient whitening}

Here,  we describe our proposal of the new extended whitening algorithms based on
\ref{sec: activation whitening}. 

In the whitening method described above,  in order to keep the approximated FIM,  $\breve{G}$,  closer to the identity matrix, updating of the whitening parameters $\{(U^{(i)},  \bm{c}^{(i)})\}$ are performed.  This makes the first factor $E\left[\bar{\bm{z}}^{\dagger^{(i-1)}}\bar{\bm{z}}^{\dagger^{(i-1)^{T}}}\right]$ in 
\begin{eqnarray}
\tilde{G}_{i,  i} &=& E\left[\bar{\bm{z}}^{\dagger^{(i-1)}}\bar{\bm{z}}^{\dagger^{(i-1)^{T}}}\right] \otimes
 E\left[\bm{\delta}^{(i)}\bm{\delta}^{(i)^{T}}\right]
\end{eqnarray}
closer to the identity matrix. 

The main idea of our method is to make the second factor $E\left[\bm{\delta}^{(i)}\bm{\delta}^{(i)^{T}}\right]$ toward the identity as well,  so that $\tilde{G}_{i,  i}$ is even better approximated by the identity matrix.  This turns out that we implement whitening process not only in the feed-forward phase but also in the back-propagating phase.

\subsubsection{Bidirectional Whitened Neural Networks} 

In order to perform the back-whitening,  we modify the forward-whitening process described by \eqref{eq: whitening forward prop1}, \eqref{eq: whitening forward prop2} and \eqref{eq: whitening forward prop3} into the following.  
\begin{eqnarray}
\label{eq: biwhitening forward prop1}
\bm{z}^{\dagger^{(i-1)}} &=& U^{(i-1)}(\bm{z}^{(i-1)} - \bm{c}^{(i-1)}) \\
\label{eq: biwhitening forward prop2}
\bm{a}^{\dagger^{(i)}} &=& W^{\dagger^{(i)}} \bm{z}^{\dagger^{(i-1)}} + \bm{b}^{\dagger^{(i)}} \\
\label{eq: biwhitening forward prop3}
\bm{a}^{(i)} &=& R^{(i)^{T}} \bm{a}^{\dagger^{(i)}} \\
\label{eq: biwhitening forward prop4}
\bm{z}^{(i)} &=& \bm{\phi}^{(i)} \left( \bm{a}^{(i)} \right)
\end{eqnarray}
Here,  $\{R^{(i)^{T}}\}$ is a newly introduced parameter,  called the back-whitening parameter. 

\begin{figure}[h]
 \centering
\includegraphics[width=10cm]{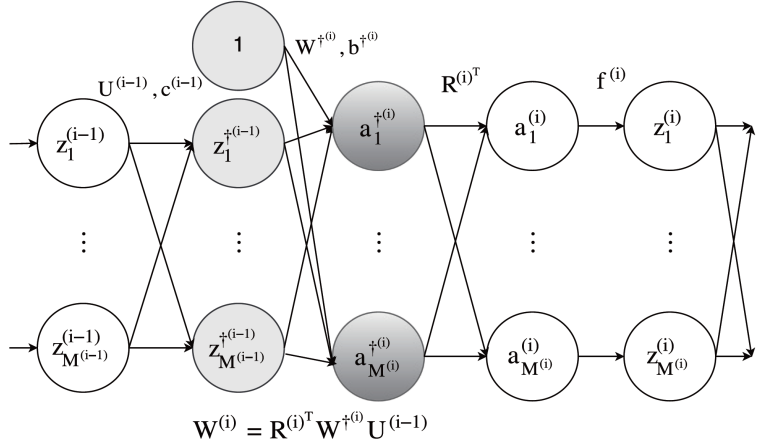}
 \caption{Architecture of the Bidirectional Whitened Neural Networks}
 \label{fig:Gradient_Whitening_Neural_Networks}
\end{figure}

We show,  as in Figure \ref{fig:Whitening Neural Networks},  the architecture of this extended method defined by \eqref{eq: biwhitening forward prop1}, \eqref{eq: biwhitening forward prop2},  \eqref{eq: biwhitening forward prop3},  \eqref{eq: biwhitening forward prop4} in the Figure \ref{fig:Gradient_Whitening_Neural_Networks}.
The dark gray part in the Figure \ref{fig:Gradient_Whitening_Neural_Networks} is the
newly introduced layer to accommodate the back-whitening parameter $\{R^{(i)^{T}}\}$. 

As mentioned above,  this proposed method performs whitening process both in feed-forward and
back-propagating phase.  Thus,  we call this new architecture as the Bidirectional Whitened Neural Networks (BWNN).

We introduce a new parameter $\bm{\delta}^{\dagger^{(i)}}$ in place of $\bm{\delta}^{(i)}$ as in the
following. 
\begin{eqnarray}
\label{eq: transformed delta}
\bm{\delta}^{\dagger^{(i)}} \equiv \dfrac{\partial l}{\partial \bm{a}^{\dagger^{(i)}}}
\end{eqnarray}
Then,  the approximation of $\tilde{G}_{i,  i}$ is then expressed as
\begin{eqnarray}
\tilde{G}_{i,  i} = E\left[  \bar{\bm{z}}^{\dagger^{(i-1)}}\bar{\bm{z}}^{\dagger^{(i-1)^{T}}}  \right]  \otimes 
 E\left[  \bm{\delta}^{\dagger^{(i)}}\bm{\delta}^{\dagger^{(i)^{T}}}  \right]
\end{eqnarray}

In analogy with Section \ref{sec: activation whitening},  we will fix the 
back-whitening parameter $\{R^{(i)^{T}}\}$ so that
\begin{eqnarray}
\label{eq: gradient whitening condition}
E\left[\bm{\delta}^{\dagger^{(i)}}\bm{\delta}^{\dagger^{(i)^{T}}}\right] = I
\end{eqnarray}

\subsubsection{Updating of the back-whitening parameter} 

Let us explicitly find $\{R^{(i)^{T}}\}$ to satisfy \eqref{eq: gradient whitening condition}.
From \eqref{eq: transformed delta},  we have
\begin{eqnarray}
\label{eq: delta relation}
\delta_{j}^{\dagger^{(i)}} = \sum_{k} \dfrac{\partial l}{\partial a_{k}^{(i)}} \dfrac{\partial a_{k}^{(i)}}{\partial a_{j}^{\dagger^{(i)}}} = \sum_{k} \delta_{k}^{(i)} r_{kj}^{(i)^{T}}
\end{eqnarray}
Thus,  $\bm{\delta}^{\dagger^{(i)}}$ is a linear transformation of $\bm{\delta}^{(i)}$,  which can be 
written as
\begin{eqnarray}
\label{eq: delta relation2}
\bm{\delta}^{\dagger^{(i)}} &=& R^{(i)}\bm{\delta}^{(i)}
\end{eqnarray}
By inserting \eqref{eq: delta relation2} into \eqref{eq: gradient whitening condition},  we obtain 
\begin{eqnarray}
E\left[\bm{\delta}^{\dagger^{(i)}}\bm{\delta}^{\dagger^{(i)^{T}}}\right]
= R^{(i)}D_{i,  i}R^{(i)^{T}} = I
\end{eqnarray}
Hence,  in analogy with \eqref{eq: update U},  $R^{(i)}$ which satisfies \eqref{eq: gradient whitening condition}
is given by the following
\begin{eqnarray}
\label{eq: update R}
R^{(i)} \leftarrow (\Lambda + \varepsilon I)^{-\frac{1}{2}} \cdot  P^{T}
\end{eqnarray}
Here,  $\Lambda,  P$ are the diagonalized and the orthogonal matrices associated with $D_{i,  i}$,  and $\varepsilon$ is
the small positive parameter to avoid a division by zero. 

Altogether,  as in the case of the forward-whitening parameters,  \eqref{eq: gradient whitening condition} is
satisfied by updating of the back-whitening parameters according to \eqref{eq: update R}, 
which,  in turn,  depends on the calculation of $\bm{\delta}^{(i)}$ in the back-propagating phase. 

\subsubsection{Updating of the model parameters}

As in the feed-forward phase, we update the model parameters $\{(W^{\dagger^{(i)}},  \bm{b}^{\dagger^{(i)}})\}$
so that the values of the multilayer Perceptron function are kept the same when the back-whitening parameters are updated. 

In order to achieve this,  the model parameters $\{(W^{\dagger^{(i)}},  \bm{b}^{\dagger^{(i)}})\}$ need to be updated as follows,  given the back-whitening parameters are updated from $R^{(i)}$ to $R_{new}^{(i)}$. 
\begin{eqnarray}
W^{\dagger^{(i)}} &\leftarrow& (R^{(i)^{T}}_{new})^{-1}R^{(i)^{T}}W^{\dagger^{(i)}} \\
\bm{b}^{\dagger^{(i)}} &\leftarrow& (R^{(i)^{T}}_{new})^{-1}R^{(i)^{T}}\bm{b}^{\dagger^{(i)}}
\end{eqnarray}
\vspace{2em}

We will call the above algorithm as ``Bidirectional Projected Natural Gradient Descent''(BPRONG) because it
performs whitening both in feed-forward and back-propagaing phase.  Its outline is shown in Algorithm \ref{alg: BiPRONG}. Also,  as in the forward-whitening,  we can perform the back-whitening update in a fixed intervals. 
They can both be done at the same time,  or independently.  In the following section,  we will employ the
latter method for a numerical application. 
\clearpage

\begin{algorithm}[tb]
\caption{Bidirectional Projected Natural Gradient Descent(BPRONG). }
\label{alg: BiPRONG}
\begin{algorithmic}
\STATE {\bf Input:} training set $D$,initial parameter $\bm{w}(0)$
\STATE {\bf Hyper parameters:} parameters for forward-whitening $\tau_{1},  c_{1}$,parameters for back-whitening $\tau_{2},  c_{2}$
\STATE $\bullet$ $U^{(i)} \leftarrow I; \bm{c}^{(i)} \leftarrow \bm{0}; R^{(i)^{T}} \leftarrow I; t \leftarrow 0$ 
\WHILE{ending condition not satisfied} 
\IF{$\mathrm{mod}(t,  \tau_{1}) = c_{1}$}
\STATE $\bullet$ forward-whitening (cf.  Algorithm \ref{alg: PRONG}).
\ENDIF
\IF{$\mathrm{mod}(t,  \tau_{2}) = c_{2}$}
\FOR{all layers i}
\STATE $\bullet$ Estimation of $D_{i,  i}$.  
\STATE $\bullet$ Computation of the back-whitening parameters $\{R^{(i)^{T}}_{new}\}$. 
\STATE $\bullet$ Updating the model parameters $\{(W^{\dagger^{(i)}},  \bm{b}^{\dagger^{(i)}})\}$. 
\STATE $\bullet$ Updating the back-whitening parameters $\{R^{(i)^{T}}\}$. 
\ENDFOR
\ENDIF
\STATE $\bullet$ Updating of $\{(W^{\dagger^{(i)}},  \bm{b}^{\dagger^{(i)}})\}$ by the ordinary gradient descent. 
\STATE $\bullet$ $t \leftarrow t + 1$ 
\ENDWHILE
\end{algorithmic}
\end{algorithm}

\section{Numerical Experiment}
\label{sec: numerical experiment}

In order to see the efficacy of our proposed method BPRONG in \ref{sec: gradient whitening},  we have applied it to
a problem of hand-written character (digits) recognition using the MNIST data set (http://yann.lecun.com/exdb/mnist/) and compared against three other methods: ordinary Stochastic Gradient Descent(SGD), Batch

\noindent
Normalization(BN)\cite{ioffe2015batch}, and PRONG. 
The network architecture is common to all the compared methods with 5 layers of 784-100-100-100-10 neurons from 
input to output.
Also,  common learning rate of $0. 01$ is taken and the mini-batch size is $100$. 
The training data contains $60000$ sets and the test data has $10000$.  We call updates of $600$ as $1$ epoch,  and 
plot,  at each epoch,  the training loss with the training set,  and the validation loss with the test data sets. 

\begin{figure}[h]
\includegraphics[width=10cm]{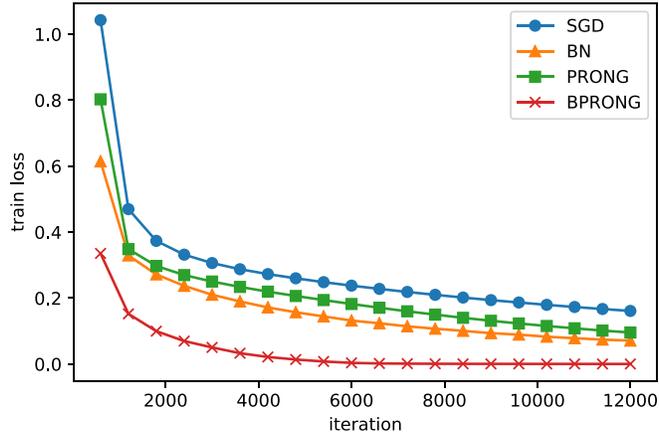}
\caption{Training loss as a function of the iteration numbers}
\label{fig: train_loss}
\end{figure}

\begin{figure}[h]
\includegraphics[width=10cm]{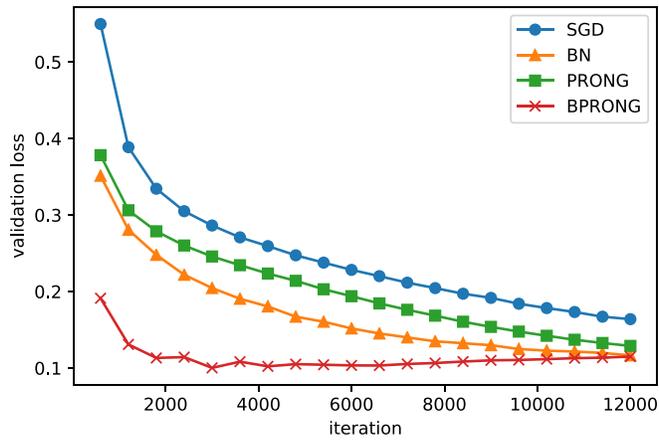}
\caption{Validation loss as a function of the iteration numbers}
\label{fig: validation_loss}
\end{figure}

\begin{figure}[h]
\includegraphics[width=10cm]{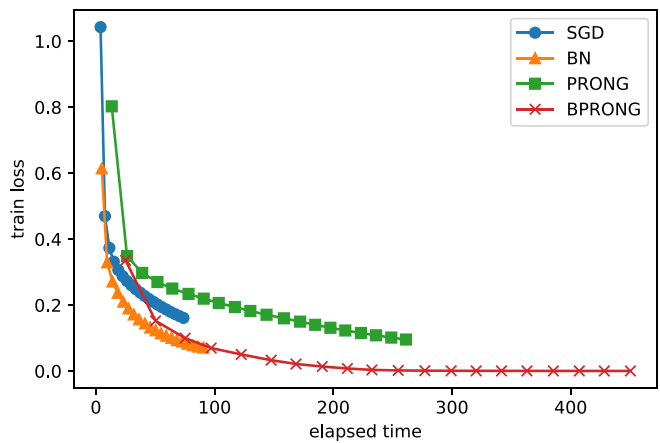}
\caption{Training loss as a function of the computational time}
\label{fig: train_loss_time}
\end{figure}

\begin{figure}[h]
\includegraphics[width=10cm]{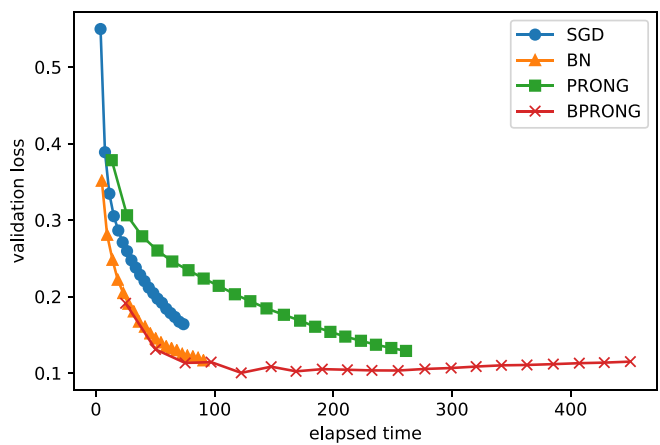}
\caption{Validation loss as a function of the computational time}
\label{fig: validation_loss_time}
\end{figure}

We observe the advantage of  BPRONG with respect to the iteration numbers both in the training and the validation
losses as shown in Figures \ref{fig: train_loss} and \ref{fig: validation_loss}.  With respect to the actual computation times,  BPRONG is faster than PRONG,  and about the same speed as  the BN (Figures \ref{fig: train_loss_time} and \ref{fig: validation_loss_time}).  This is due to the fact that eigenvalue decomposition 
associated with the whitening is computationally costly to offset the advantage over BN with respect to 
iteration numbers.  

Altogether,  our proposed method, BPRONG, has shown its potential.  If we can find methods to speed up the whitening process,  BPRONG can show its effectiveness further. 

\clearpage

\section{Discussion}
\label{sec: discussion}
We presented here an extended model of the previously proposed Whitened Neural Networks\cite{NIPS2015_5953} as a method to realize the Natural Gradient Descent. 
Our extension,  which we call Bidirectional Whitened Neural Networks,  aims to make the Fisher
Information Matrix closer to the identity matrix.  It has shown its potential as an efficient 
method thorough a numerical application to a hand-written digits recognition problem.  

We note two points as topics to be investigated further.  First,  the proposed model should be tested 
for larger and deeper network architectures for a check of its efficacy and stability.  It may
require further modifications for improvements on these aspects,  particularly by exploring matrix decomposition methods.  Secondly,  we want to find more dynamical way for whitening process.  In other words, we would like to keep the
Fisher Information Matrix constantly closer to the identity by continuous whitenings.  Though it is
computationally more expensive,  we may build on some previous studies, such as adaptive calculations of transforming matrices\cite{cardoso1996equivariant}. 



\bibliographystyle{plain}
\bibliography{bwnn}


\end{document}